\documentclass[lettersize,journal]{IEEEtran}
\usepackage{amsmath,amsfonts}
\usepackage{algorithmic}
\usepackage{algorithm}
\usepackage{array}
\usepackage[caption=false,font=normalsize,labelfont=sf,textfont=sf]{subfig}
\usepackage{textcomp}
\usepackage{stfloats}
\usepackage{url}
\usepackage{verbatim}
\usepackage{graphicx}
\usepackage{cite}
\usepackage{xcolor}
\usepackage{multirow}
\usepackage{booktabs}
\usepackage{ulem}
\usepackage{graphicx}
\usepackage{float}
\usepackage{bbding}
\usepackage{rotating}
\usepackage{subcaption}

\begin{document}

\title{ME-TST+: Micro-expression Analysis via Temporal State Transition with ROI Relationship Awareness}

\author{Zizheng Guo,~\IEEEmembership{Student Member,~IEEE,}  Bochao Zou,~\IEEEmembership{Member,~IEEE,} Junbao Zhuo, Huimin Ma,~\IEEEmembership{Senior Member,~IEEE}
\thanks{This work was supported in part by the National Natural Science Foundation of China (62206015, 62227801, 62402033), the Fundamental Research Funds for the Central Universities (FRF-KST-25-008), and the Young Scientist Program of The National New Energy Vehicle Technology Innovation Center (Xiamen Branch).(Corresponding author: Bochao Zou)}
\thanks{Z. Guo, B. Zou, J. Zhuo and H. Ma are with the School of Computer and Communication Engineering, University of Science and Technology Beijing, 100083, China (e-mail: guozizheng@xs.ustb.edu.cn; zoubochao@ustb.edu.cn; junbaozhuo@ustb.edu.cn; mhmpub@ustb.edu.cn.)}
}

\markboth{Journal of \LaTeX\ Class Files,~Vol.~14, No.~8, August~2021}%
{Shell \MakeLowercase{\textit{et al.}}: A Sample Article Using IEEEtran.cls for IEEE Journals}

\IEEEpubid{0000--0000/00\$00.00~\copyright~2021 IEEE}

\maketitle
\begin{abstract}
Micro-expressions (MEs) are regarded as important indicators of an individual's intrinsic emotions, preferences, and tendencies. ME analysis requires spotting of ME intervals within long video sequences and recognition of their corresponding emotional categories. Previous deep learning approaches commonly employ sliding-window classification networks. However, the use of fixed window lengths and hard classification presents notable limitations in practice. Furthermore, these methods typically treat ME spotting and recognition as two separate tasks, overlooking the essential relationship between them. To address these challenges, this paper proposes two state space model-based architectures, namely ME-TST and ME-TST+, which utilize temporal state transition mechanisms to replace conventional window-level classification with video-level regression. This enables a more precise characterization of the temporal dynamics of MEs and supports the modeling of MEs with varying durations. In ME-TST+, we further introduce multi-granularity ROI modeling and the slowfast Mamba framework to alleviate information loss associated with treating ME analysis as a time-series task. Additionally, we propose a synergy strategy for spotting and recognition at both the feature and result levels, leveraging their intrinsic connection to enhance overall analysis performance. Extensive experiments demonstrate that the proposed methods achieve state-of-the-art performance. The codes are available at https://github.com/zizheng-guo/ME-TST.
\end{abstract}

\begin{IEEEkeywords}
Micro-expression analysis, Synergistic spotting and recognition, Temporal
state transition, Long videos.
\end{IEEEkeywords}

\section{Introduction}
\label{sec:intro}

Facial expressions are key indicators of emotional states and can subtly communicate intentions, choices, and preferences.  Based on their intensity and duration, these expressions are generally categorized as macro-expressions~(MaEs) or micro-expressions~(MEs). Representative works on MEs~\cite{tpami-review,tac-survey,casme3} and MaEs~\cite{fer1,fer2,fer3} provide detailed investigations into these two categories. Compared to MaEs, MEs are not only characterized by lower intensity and shorter duration, but are also considered genuine reflections of emotion that are difficult to feign or suppress. They are particularly likely to manifest when individuals attempt to conceal or suppress their MaEs. Accordingly, the analysis of MEs holds significant value in domains such as medical diagnosis, as well as business and political negotiations.

ME analysis involves two fundamental tasks: spotting and recognition. Spotting pertains to accurately determining the onset and offset of MEs within lengthy video sequences, while recognition focuses on categorizing the emotions expressed during these detected intervals. The vast majority of studies have addressed these two tasks separately. In the spotting domain, there has been a discernible shift from traditional signal processing techniques to deep learning-based methods~\cite{mesnet,lgsnet}, although classic approaches continue to achieve competitive results~\cite{MEGC2023}. By contrast, research on recognition has been more active~\cite{recog1,recog2,recog_bert}, with deep learning methods emerging as the dominant paradigm in this field.

\IEEEpubidadjcol

Despite considerable progress in the respective sub-tasks, few studies have directly addressed the unified task of ME analysis—that is, jointly spotting and recognizing emotions from long video sequences, which better reflects real-world application scenarios. Whether by integrating separate spotting and recognition frameworks or employing a unified network, current approaches generally fall short in this respect.
Traditional signal processing methods~\cite{li2017,Liong2017_MA} extract handcrafted features for spotting and recognition, but are generally limited to short video segments containing a single ME and do not account for MaEs. More recent studies~\cite{MEAN,sfamnet,megc24-1,megc24-2} leverage deep learning techniques to sequentially perform spotting followed by recognition within long video streams. However, these methods predominantly rely on classification networks combined with sliding window mechanisms, where fixed window sizes and hard window-level classification introduce intrinsic constraints. Such designs often struggle to capture the full diversity and temporal dynamics of MEs, thereby limiting overall performance. Furthermore, these frameworks continue to treat spotting and recognition as two independent tasks, failing to fully exploit the complementary nature of their relationship.

\begin{figure}[t] 
  \centering 
  \includegraphics[width=0.99\linewidth]{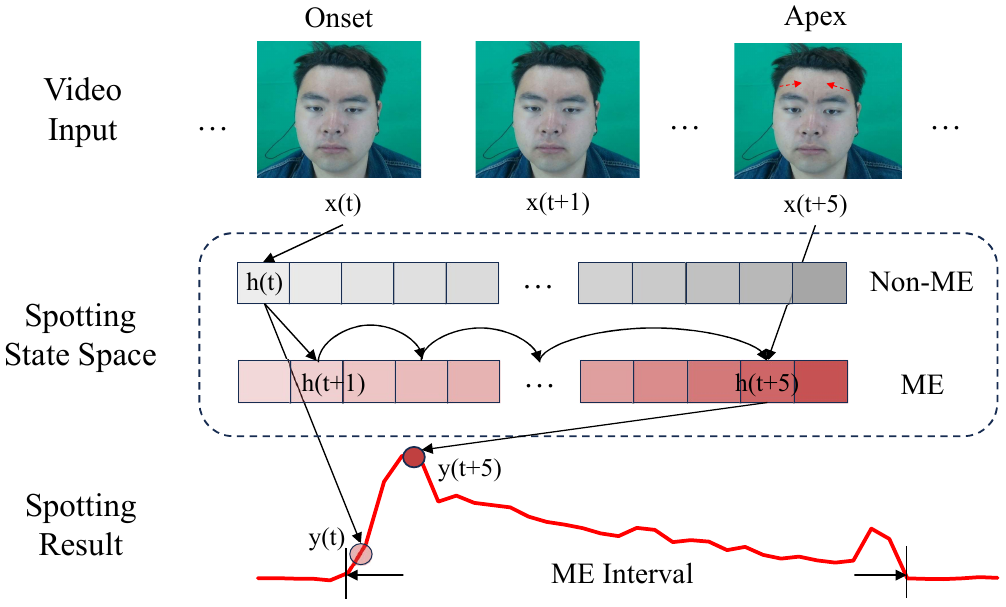}  
  \caption{A schematic diagram of state transitions. A non-ME state, $h(t)$, transitions to an ME state, $h(t+1)$, upon receiving the onset frame input, $x(t)$, resulting in output $y(t)$. As subsequent frames are processed, the intensity of the ME state gradually increases, culminating at the apex frame.}
  \label{fig:1}
\end{figure}

To address these challenges, we propose a novel ME analysis method based on temporal state transitions, in which video-level regression replaces the conventional window-level classification framework. This approach enables flexible handling of MEs of varying durations and facilitates comprehensive modeling of their temporal dynamics. As depicted in Fig.~\ref{fig:1}, the temporal progression of ME phases—including onset, apex, and offset—is represented as state transitions within a defined state space.
Furthermore, we introduce a synergistic spotting and recognition strategy that equips the recognition branch with partial spotting capability. At the feature level, this design enables spotting and recognition to align within a shared module during training, thereby fully leveraging limited data to uncover low-level features. At the result level, the spotting capacity of the recognition pathway assists in identifying challenging scenarios, such as distinguishing blinks and other ambiguous intervals. By exploiting the complementary strengths of both branches in terms of feature representation and output results, the proposed strategy substantially improves overall performance.

A preliminary version of this work appeared in~\cite{METST}. Compared to the previous version, the main advancements in this work are as follows:
(1) The preprocessing approach and network design in the prior work resulted in substantial spatial information loss. In this paper, we introduce multi-granularity ROI modeling and adopt the slowfast architecture to enhance spatial information extraction, thereby more effectively demonstrating the potential of the proposed temporal state transition architecture.
(2) The methodology is described in greater detail, accompanied by more extensive experimental results and thorough analyses.
(3) Extensive visualizations are included, offering further insights into the working mechanism of the proposed new architecture.

The main contributions can be summarized as follows:

$\bullet$ We propose a temporal state transition architecture, which replaces window-level classification with video-level regression. To the best of our knowledge, this is the first work to investigate state space models in the ME domain.

$\bullet$ We design a synergy strategy at both the feature and result levels to exploit the inherent connections between spotting and recognition, thereby optimizing overall analysis performance.

$\bullet$ We conduct extensive experiments demonstrating that the proposed method achieves state-of-the-art performance. Furthermore, a comprehensive set of visualizations and data analyses is provided to illustrate the effectiveness of the proposed architecture.

\section{Related Work}
\label{sec:relatedwork}

\subsection{ME Spotting}
Early research on ME spotting primarily relied on handcrafted feature extraction methods, which can be broadly categorized into two groups: optical flow-based methods and feature descriptor-based methods. Optical flow-based approaches include strain pattern analysis~\cite{review36,review73,review74}, Discriminative Response Map Fitting (DRMF) models~\cite{review75,review102}, and the Main Direction Maximum Difference (MDMD) method~\cite{review77}. Feature descriptor methods encompass the gradient histogram descriptor~\cite{review23,review78}, Local Binary Patterns (LBP)~\cite{review79}, Histogram of Oriented Gradients (HOG)~\cite{cvpr25-4}, SIFT descriptors~\cite{review80}, and Constrained Local Models (CLM)~\cite{li2017,review105,review106}. Overall, traditional approaches are hampered by high computational costs, pronounced sensitivity to noise, and heavy reliance on parameter tuning~\cite{tpami-review}.

With the advent of machine learning, works such as~\cite{casme38} and~\cite{casme42} employed SVM classifiers for ME frame identification. Subsequently, the rise of deep learning has shifted the research paradigm from traditional "feature engineering" to "end-to-end learning." Rather than simply substituting handcrafted features, these features are frequently incorporated into neural networks as input representations, as illustrated by studies such as~\cite{lssnet,softnet}. Building on this foundation, numerous studies have sought to further optimize network architectures, including the application of 3D-CNN~\cite{3D-CNN} and (2+1)D-CNN architectures~\cite{mesnet}. In addition, Yang et al.~\cite{yang2021facial} constructed an end-to-end framework that incorporates deep features derived from facial action units, while Leng et al.~\cite{abpn} proposed a vertex- and boundary-aware network designed to mitigate head movement artifacts. Collectively, these methods synergistically combine the nonlinear feature learning capacity of deep neural networks with the physical interpretability of traditional features, thereby enhancing both robustness and accuracy.
From the perspective of video sequence processing, these deep learning methods can be further classified into key-frame-based approaches~\cite{casme39,weakly36,weakly37,weakly38,weakly39} and interval-based approaches~\cite{weakly40,weakly41,weakly42}. Key-frame-based methods detect MEs by identifying a single representative frame, typically the apex; however, such approaches may fail to capture the full temporal dynamics of MEs. In contrast, interval-based methods exploit temporal information from adjacent frames to more accurately determine the ME intervals. Despite notable progress in recent years, critical challenges in ME spotting remain unresolved, and deep learning-based techniques have yet to consistently surpass traditional approaches in performance~\cite{MEGC2023}.

\subsection{ME Recognition}
Early methods for ME recognition predominantly relied on handcrafted feature extraction, aiming to capture the subtle and transient movements of facial muscles. A representative example is Local Binary Patterns from Three Orthogonal Planes (LBP-TOP)~\cite{tpami25-7,cvpr25-26}, which effectively integrates spatiotemporal information and laid the groundwork for subsequent feature development. Building on this, researchers introduced techniques such as Histogram of Oriented Gradients (HOG)~\cite{cvpr25-4}, Histogram of Image Gradient Orientations (HIGO)~\cite{li2017}, Fuzzy Histogram of Oriented Optical Flow (FHOOF)~\cite{tpami25-6,tpami25-16}, and Bi-Weighted Oriented Optical Flow (Bi-WOOF)~\cite{casme39}, continuously enhancing the accuracy of motion representation for MEs. Nevertheless, despite advancements achieved by these handcrafted-feature-based methods, they tend to be sensitive to noise and heavily dependent on the accurate extraction of key video frames, which hinders their generalization and robustness in real-world scenarios.

In recent years, deep learning approaches have attracted increasing attention. End-to-end deep neural networks are capable of automatically extracting more discriminative spatiotemporal features from raw data, thereby overcoming the limitations of manually designed features. Many studies combine optical flow and other features with deep models; for instance, methods such as OFF-ApexNet~\cite{off-apexnet} and STSTNet~\cite{ststnet} utilize multi-stream architectures to separately process different components of optical flow, thus improving the modeling of dynamic ME features. Subsequently, several works introduced various auxiliary priors and designs, leveraging innovative networks to obtain more effective feature representations and further enhance performance~\cite{tpami25-21,tpami25-13,tpami25-18,mfdan_tcvst,soda4mer,mol}. Although significant progress has been made, most mainstream MER paradigms still rely on predefined ME intervals, requiring precise annotation of key frames and the use of frame differences between onset and apex frames. This limitation impedes the direct automated analysis of long video sequences and considerably restricts the widespread application of ME recognition in real-world scenarios.

\subsection{ME Analysis}
ME analysis can be regarded as an integrated task encompassing both ME spotting and recognition, which requires the localization of ME intervals and the identification of corresponding emotions within long video sequences. This task closely aligns with the requirements of real-world application scenarios.
Although ME analysis is highly relevant to real-world applications, research in this area has significantly lagged behind studies focusing on its two subtasks individually, and its overall progress remains limited. Earlier works, such as those by Li et al.~\cite{li2017} and Liong et al.~\cite{Liong2017_MA}, adopted traditional signal processing approaches that extract handcrafted features for either spotting or recognition. However, these methods are generally constrained to short video clips containing a single ME instance and overlook MaEs, which restricts their applicability in practical scenarios. In recent years, approaches like MEAN~\cite{MEAN} and SFAMNet~\cite{sfamnet} have utilized deep learning architectures to perform sequential ME spotting and recognition in longer videos. Within the MEGC2024 challenge, further advances have been made: USTC-IAT-United employed the VideoMAE framework to more effectively model the temporal dynamics of MEs~\cite{megc24-1}, while He Yuhong et al.~\cite{megc24-2} addressed class imbalance through targeted data preparation and proposed a three-phase pipeline involving ME spotting, emotion recognition, and the exclusion of non-ME segments.

Although these studies have substantially advanced the field of ME analysis, they remain largely confined to frameworks that treat spotting and recognition as distinct subtasks. Most existing approaches employ classification networks with sliding window strategies, handling these two processes independently. In contrast, we propose a novel temporal state transition architecture that synergistically integrates ME spotting and recognition.

\subsection{State Space Model}
To address the challenges associated with modeling long sequences, state space models (SSMs) have attracted substantial scholarly interest. Functionally, SSMs can be considered hybrids between recurrent neural networks (RNNs) and convolutional neural networks (CNNs). Notable in this line of work is the structured state space model (S4)~\cite{s4}, along with its subsequent variants~\cite{s5,h3,gss}.

Among recent advancements, Mamba~\cite{mamba,mamba2} stands out for its introduction of a data-dependent SSM, coupled with a parallel scan-based selection mechanism that achieves linear computational complexity without sacrificing the ability to model long-range dependencies. In contrast to Transformer architectures, whose attention mechanism entails quadratic complexity~\cite{transformer,vivit}, Mamba demonstrates particular strengths in efficiently processing long sequences. This promising architecture has inspired a range of follow-up studies~\cite{visionmamba,simba,videomamba}, which collectively report that Mamba outperforms Transformer-based models in both predictive accuracy and GPU computational efficiency across various vision tasks.

Nevertheless, it has been pointed out in the literature that when spatial information is introduced as an additional sequence dimension in visual tasks, the dimensionality of the state transitions increases substantially. Such an expansion can undermine Mamba’s ability to effectively model temporal dynamics, particularly in scenarios where the target signals are weak or subtle~\cite{rhythmmamba}. ME analysis serves as a prime example: MEs are characterized by extremely subtle facial muscle movements of very short duration, making them highly susceptible to noise from variations in illumination and inadvertent movements. Experimental results indicate that simply substituting the Transformer block with a vanilla Mamba block in ME analysis tasks results in significant performance degradation.

Departing from conventional approaches, we reframe ME analysis as a time-series task. This perspective offers several key advantages. First, from an SSM standpoint, it reduces the dimensionality of state transitions by constraining them exclusively to the temporal axis. This matches the inherent phase structure of MEs—onset, apex, and offset—enabling more precise modeling of these transient events. Second, from the ME analysis perspective, this formulation allows the model to focus on the dynamic evolution of facial movements over time, rather than being distracted by spatial redundancies. By isolating temporal dependencies, the model is better equipped to spot and recognize subtle MEs, which are typically characterized by brief and barely perceptible changes. This focus is especially beneficial for mitigating the impact of noise commonly encountered in real-world scenarios, such as illumination fluctuations and inadvertent head movements.
\section{Methodology}
\label{sec/3_methodology} 

\begin{figure*}[t]
  \centering  
  \includegraphics[width=0.98\linewidth]{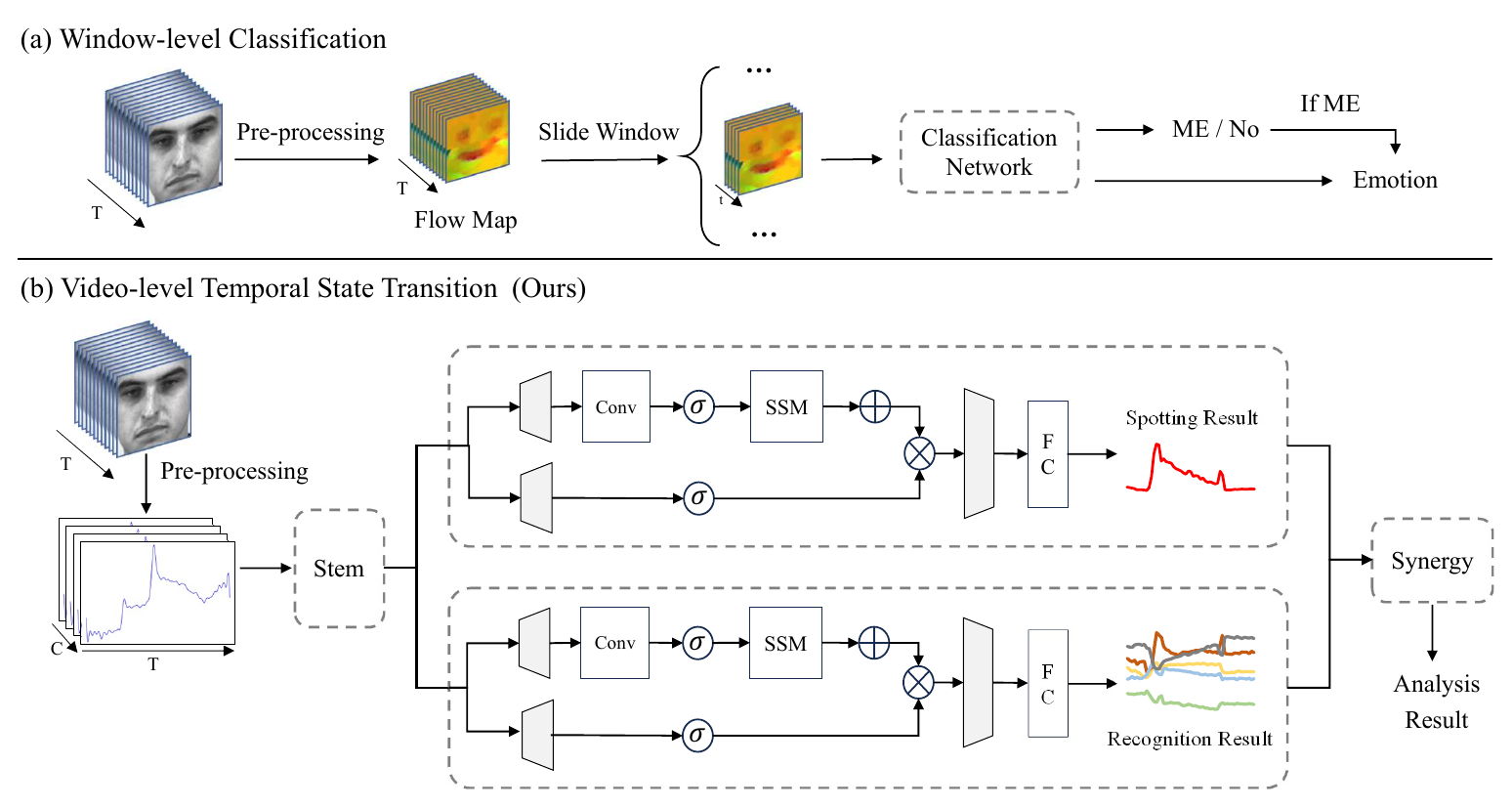}  
  \caption{(a) The framework of the window-level classification method. (b) The framework of the proposed ME-TST. Where "$+$" represents addition, "$\times$" represents multiplication, "$\sigma$" represents the activation layer, and the trapezoid represents the linear layer.}
  \label{fig:main}
\end{figure*}

We will first introduce the architecture of ME-TST and ME-TST+ in Sec. 3.1 and 3.2,
respectively. And at last present the synergy strategy in Sec. 3.3.

\begin{figure*}[t]
  \centering  
  \includegraphics[width=0.98\linewidth]{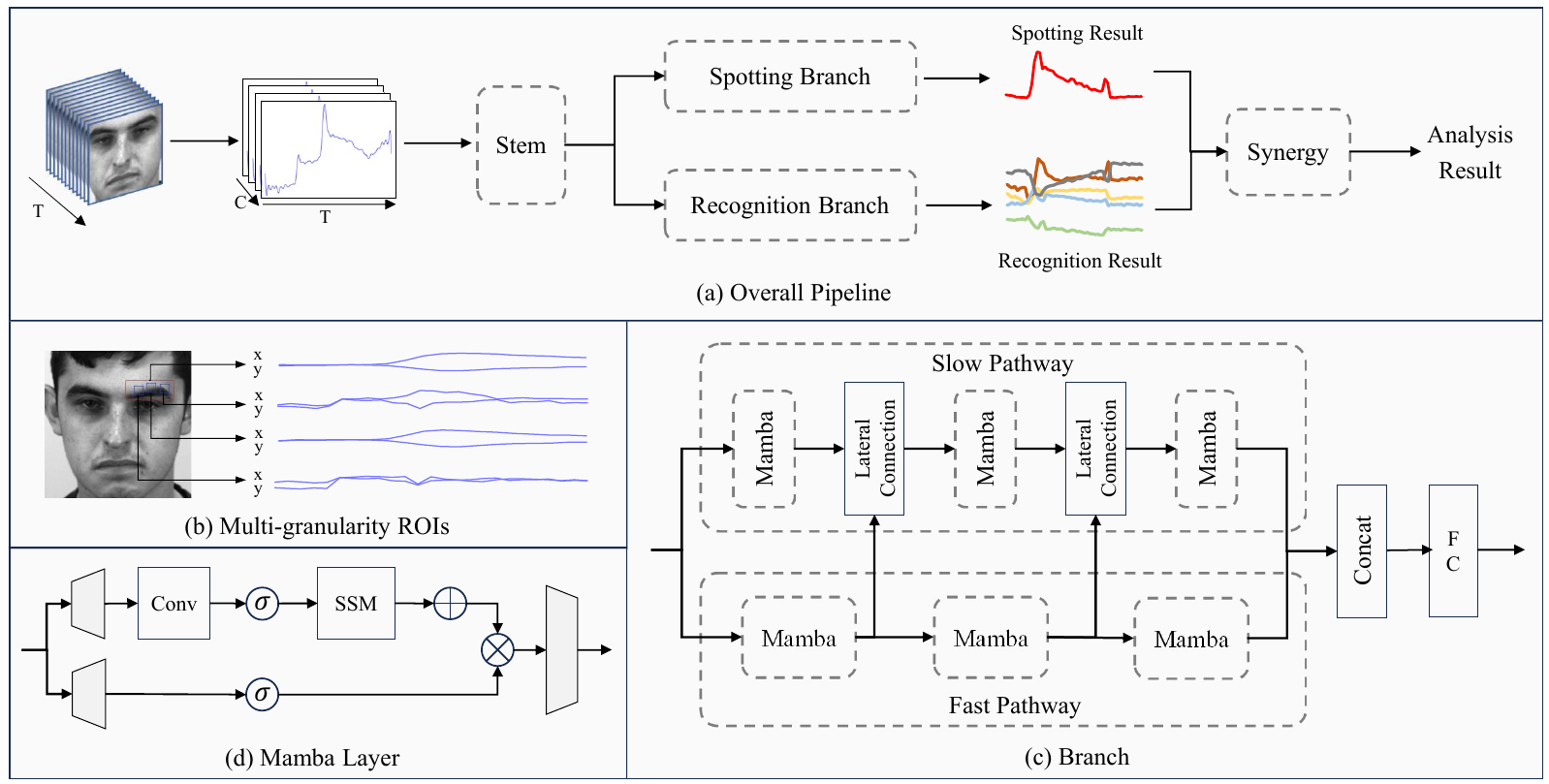}  
  \caption{Detailed architecture of the proposed ME-TST+ framework.}
  \label{fig:2}
\end{figure*}

\subsection{ME-TST} 
The overall structure of the proposed approach is depicted in Fig.~\ref{fig:main}~(b). For each input facial video, a pre-processing stage first extracts optical flow from predefined regions of interest (ROIs), as outlined in prior works~\cite{MEGC2023-2, he-megc2021-1}. Optical flow is extracted using a sliding window with overlap, and the optical flow from each window is aggregated to construct video-level optical flow representations. Compared to conventional frame-by-frame optical flow accumulation, this window-based approach more effectively captures temporal changes in MEs over segments, enables smooth tracking across multiple time intervals, and mitigates the impact of accumulated frame-wise errors. Each optical flow vector comprises both x and y components, resulting in an inter-frame optical flow sequence for the ROIs, denoted by $X_{flow} \in \mathbb{R}^{C_1\times T}$, where $C_1$ is the number of channels and T denotes the number of video frames. To ensure consistency, each frame is further aligned using global optical flow from the overall facial region.

Subsequently, the optical flow sequences undergo initial feature extraction in the stem module, which consists of two sequential 1D convolutional layers, each followed by batch normalization and ReLU activation functions. The resultant features are then propagated into two parallel pathways: one dedicated to ME spotting and the other to ME recognition. These branches share identical architectures based on state-space models, but maintain separate, non-shared weights. By modeling the input as a one-dimensional time series, the state transitions within the state-space model can explicitly capture the temporal dynamics associated with MEs. Cross-channel projections in both paths facilitate interaction between channels, thereby enhancing the learning of relationships between ROI motion patterns and ME categories.

The SSM, a form of linear time-invariant system, maps inputs to outputs via latent states, and is described by the following ordinary differential equations:
\begin{equation}
\begin{split}
&h'(t)=Ah(t)+Bx(t),\\
&y(t)=Ch(t).
\end{split}
\end{equation}

Where $A\in \mathbb{R}^{N\times N}$ is the evolution matrix, $B\in \mathbb{R}^{N\times 1}$ and $C\in \mathbb{R}^{1\times N}$ are the input and output projection matrices, and $N$ is the sequence length. Practical challenges arise from directly applying continuous SSMs; thus, the Mamba framework adopts a discrete formulation, mapping A and B to discrete counterparts $\overline{A}$ and $\overline{B}$ with a time-scale parameter $\Delta$, typically discretized via the zero-order hold method:
\begin{equation}
\begin{split}
&\overline{A} = \exp(\Delta A),\\
&\overline{B} = (\Delta A)^{-1}(\exp(\Delta A) - I)\Delta B,\\
&h_t = \overline{A}h_{t-1} + \overline{B}x_t,\\
&y_t = Ch_t.
\end{split}
\end{equation}

However, due to the intrinsic sequential dependencies of this formulation, efficient implementation relies on expressing it as a convolution:
\begin{equation}
\begin{split}
&\overline{K} = (C\overline{B}, C\overline{A}\overline{B}, \dots, C\overline{A}^{N-1}\overline{B}),\\
&y = x \ast \overline{K},
\end{split}
\end{equation}

Where $\overline{K}\in \mathbb{R}^N$ denotes a structured convolution kernel, and $\ast$ indicates a convolution operation.
The two SSM-based branches generate their respective outputs: $X_{spot} \in \mathbb{R}^{T\times 1}$ for ME spotting, and $X_{recog} \in \mathbb{R}^{T \times (emo+1)}$ for ME recognition, where emo corresponds to the number of emotion classes, with an additional class indicating "neutral"~(non-ME).

Finally, the outputs from ME spotting and recognition are integrated during post-processing. Peak detection is applied to the spotting results to identify the temporal intervals corresponding to MEs, and the predominant emotion within each interval is determined from the recognition outputs. In this way, the final results—comprising both the temporal interval and the associated emotion of each ME—are obtained.

\subsection{ME-TST+}
Unlike previous work, we approach ME analysis as a time-series problem rather than the conventional video understanding task, leveraging video-level optical flow extraction from key ROIs. In ME-TST, using only five coarse-grained ROIs for optical flow extraction leads to considerable loss of spatiotemporal information. This rather simplistic strategy often results in insufficient characterization, making it difficult to accurately capture the diverse and subtle ME patterns that differ across individuals. Consequently, it significantly limits the representational capacity of subsequent models and increases the risk of overfitting. In addition, ME-TST uses the fixed token length, which may be suboptimal for MEs with varying temporal spans. Larger token lengths reduce temporal redundancy but also diminish fine-grained temporal cues, while smaller token lengths have the opposite effect. 

To address these limitations, we propose a temporally enhanced ME-TST+ (see Fig.~\ref{fig:2}), which comprehensively extracts spatial information from key ROIs at multiple granularity and analyzes it using a dual-path slowfast network with both long and short temporal windows. The fast pathway samples the input sequence at a high frame rate, enabling efficient encoding of fine-grained temporal cues during ME dynamics and effectively capturing subtle facial movements within short time spans. In contrast, the slow pathway processes spatially enhanced features at a lower temporal resolution, focusing on modeling long-term semantic motion patterns. This SlowFast architecture, through a cross-path interaction mechanism, adaptively fuses multi-scale spatiotemporal contextual information, significantly enhancing the perception of weak and transient MEs, thereby improving the robustness and accuracy of ME analysis.

Specifically, as shown in Fig.~\ref{fig:2}(c), the input features are fed in parallel into the slow pathway and the fast pathway. The fast pathway receives features at the original temporal resolution, denoted as $X_{fast} \in \mathbb{R}^{C\times T}$, while the slow pathway processes a temporally downsampled sequence with input $X_{slow} \in \mathbb{R}^{2C\times T/2}$, featuring a higher channel dimension to preserve richer spatial semantic information. Subsequently, both pathways independently extract features through three cascaded Mamba Blocks, each consisting of four stacked Mamba layers that effectively capture long-range temporal dependencies, leveraging their state-space modeling capabilities. To facilitate cross-path information interaction, lateral connections are introduced after each Mamba Block, expressed as $X_{slow} = X_{slow} + Conv(X_{fast})$. Through learnable convolutional transformations, mid-level fine-grained dynamic cues from the fast pathway are adaptively injected into the slow pathway. This not only supplements temporal detail information in the slow pathway but also mitigates the loss of ME features caused by temporal downsampling, enhancing the sensitivity of the slow pathway to instantaneous changes and improving the accuracy of critical temporal points such as onset, apex, and offset. Finally, high-level semantic features from both pathways are concatenated along the channel dimension and passed through a fully connected layer to produce the final prediction.

\begin{figure}[t]
  \centering  
  \includegraphics[width=0.95\linewidth]{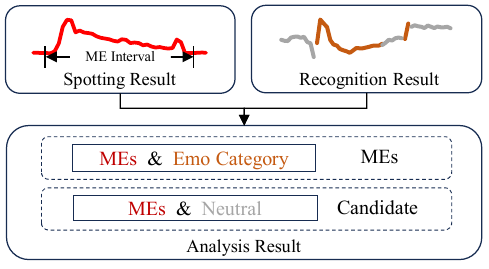}  
  \caption{The schematic diagram of the result-level synergy strategy.}
  \label{fig:3}
\end{figure}

\subsection{Synergistic Spotting and Recognition}
In previous research on ME analysis, the predominant paradigm has been “spot then recognize,” in which the recognition stage is performed based on intervals identified by the spotting module. Under this framework, it is assumed that all instances requiring recognition are MEs; accordingly, the neutral (non-ME) category is typically not distinguished. Even when included for annotation convenience, the loss weight for the neutral category is often set to zero during training~\cite{MEAN,sfamnet}. This design hampers the alignment of spotting and recognition during training, thereby severely limiting the potential for shared feature representations and joint optimization.

The core of our proposed collaborative spotting and recognition strategy is the explicit integration of the neutral category into the training process, which effectively allocates part of the spotting capability to the recognition branch. This integration fosters synergy between the spotting and recognition branches, comprising two components: feature-level synergy and result-level synergy.
For feature-level synergy, the spotting and recognition branches share a stem module for primary feature extraction. The integration of spotting capability within the recognition branch promotes the alignment of features leveraged by both tasks, thus facilitating more effective low-level representation learning and enhancing overall performance.
For result-level synergy, because the recognition branch acquires a certain degree of spotting ability, the final result analysis can incorporate recognition results to assist in spotting. For instance, while spotting eye blinks remains a challenge for the spotting branch, they may be more easily detected by recognition. As illustrated in Fig.~\ref{fig:3}, cases in which the spotting branch predicts ME while the recognition branch yields a neutral result are considered as ME candidates. Intuitively, the number of false positives (FPs) among ME candidates greatly exceeds the number of true positives (TPs). Thus, classifying these instances as non-ME may incur a slight loss in recall but results in a substantial improvement in overall performance. This intuition is substantiated by our ablation experiments.
\label{sec/4_experiment}
\section{Experiment}

\subsection{Dataset and Performance Metric}
\textbf{Datasets}. Two long-video ME benchmarks are utilized for evaluation: CAS(ME)$^3$~\cite{casme3} and SAMMLV~\cite{samm,SAMMLV}. CAS(ME)$^3$ contains 1,300 long videos from 100 subjects, recorded at 30 frames per second (fps), with annotations for 3,342 MaEs and 860 MEs. SAMMLV comprises 147 long videos from 32 subjects at 200 fps, annotated with 343 MaEs and 159 MEs; only MEs are provided with emotion category labels.

\textbf{Performance Metric}. Subtasks of ME spotting and recognition are assessed using respective task-specific metrics, along with a composite metric quantifying overall performance.
For the spotting task, F1-score and recall are reported, adopting the Intersection-over-Union (IoU) matching criterion as described in~\cite{iou}:
\begin{equation}
\mathrm{IoU} = \frac{| I_{\mathrm{pred}} \cap I_{\mathrm{gt}} |}{| I_{\mathrm{pred}} \cup I_{\mathrm{gt}} |}.
\end{equation}

Where $I_{pred}$ and $I_{gt}$ denote the predicted and ground-truth time intervals, respectively. A detected interval is considered a true positive (TP) if its IoU with the ground-truth interval exceeds 0.5.
For the recognition task, F1-score, unweighted average recall (UAR), and unweighted F1-score (UF1)~\cite{uf1uar} are reported. The evaluation metrics are defined as follows:
\begin{align}
\mathrm{UAR} &= \frac{1}{C} \sum_{c=1}^{C} \frac{\mathrm{TP}_c}{\mathrm{TP}_c + \mathrm{FN}_c}, \\
\mathrm{UF1} &= \frac{1}{C} \sum_{c=1}^{C} \frac{2\,\mathrm{TP}_c}{2\,\mathrm{TP}_c + \mathrm{FP}_c + \mathrm{FN}_c}.
\end{align}

Where C denotes the number of classes, $TP_c$, $FP_c$, and $FN_c$ represent the numbers of true positives, false positives, and false negatives for class c, respectively.
Overall analysis is conducted using the Spot-Then-Recognize Score (STRS)~\cite{MEAN} as a composite metric:
\begin{equation}
\mathrm{STRS} = F1_\mathrm{spotting} \times F1_\mathrm{recognition}.
\end{equation}

\subsection{Implementation Details}
For facial video input, optical flow is extracted from 18 ROIs at different granularities using the Gunnar Farnebäck algorithm~\cite{flow}. Each optical flow vector is represented by its x and y components. Consequently, the resulting feature dimension $C_1$ is 36. This feature dimension is then increased to 128 by the main network. After feature extraction, the outputs from the slow and fast pathways are concatenated, resulting in a combined channel dimension $C_2$ of 384.

The spotting branch employs Mean Squared Error (MSE) as the loss function, formally defined as:
\begin{equation}
\mathcal{L}_{\mathrm{spot}} = \frac{1}{N} \sum_{i=1}^N (y_i - \hat{y}_i)^2.
\end{equation}

Where $y_i$ and $\hat{y}_i$ denote the ground-truth and predicted spotting scores for the $i_{th}$ sample, respectively, and N is the number of samples.
For the recognition branch, we employ the cross-entropy loss, formulated as:
\begin{equation}
\mathcal{L}_{\mathrm{recog}} = -\sum_{i=1}^{N} \sum_{c=1}^{C} w_c, y_{ic} \log{\hat{y}_{ic}}.
\end{equation}

Where $w_c$ denotes the class-specific weight for class c; C is the number of emotion classes, including neutral; and $y_{ic}$ and $\hat{y}_{ic}$ denote the groundtruth and predicted probabilities for the $i_{th}$ sample belonging to class c, respectively.
Model performance was rigorously evaluated using the Leave-One-Subject-Out (LOSO) cross-validation protocol. The network was trained for 50 epochs with an initial learning rate of 3e-4. All training procedures were executed on a single NVIDIA RTX 4090 GPU.

\begin{table}[t]
  \caption{ME Spotting Evaluation~(F1-Score~$\uparrow$).}
  \centering
  \setlength{\tabcolsep}{6.5pt}
    \begin{tabular}{cccc}
    \toprule
    Type & Method & CAS(ME)$^3$ & SAMMLV \\
    \midrule
    \multirow{16}[2]{*}{S}           
          & MDMD~\cite{he}  & -     & 0.0364 \\
          & SP-FD~\cite{zhang2020} & 0.0103 & 0.1331 \\
          & OF-FD~\cite{he-megc2021-1} & 0.0000 & 0.2162 \\
          & MESNet~\cite{mesnet} & - & 0.0880 \\
          & SOFTNet~\cite{softnet}  & - & 0.1520 \\
          & LSSNet~\cite{lssnet} & 0.0653 & 0.2180 \\
          & ABPN~\cite{abpn} & - & 0.1689 \\
          & MTSN~\cite{mtsn} & - & 0.0878 \\
          & Advanced Concat-CNN~\cite{AdvancedConcatCNN} & - & 0.2282 \\
          & LGSNet~\cite{lgsnet} & \textbf{0.0990} & - \\
          & Yang et al.~\cite{yang2021facial} & - & 0.1155 \\
          & Li et al.~\cite{li2024learning} & - & 0.2541 \\
          & WCMN~\cite{zhou2024micro} & - & 0.2330 \\
          & Ofct~\cite{ofct} & - & 0.2466 \\
          & MSOF~\cite{msof} & - & 0.2457 \\
          & Causal-Ex~\cite{Causal-Ex} & - & 0.2010 \\
          & MC-WES~\cite{yu2025weakly} & 0.0000 & 0.1350 \\
    \midrule
    \multirow{6}[2]{*}{S \& R} & MEAN~\cite{MEAN}  & 0.0283 & 0.0949 \\
          & SFAMNet~\cite{sfamnet}& 0.0716 & - \\
          & USTC-IAT-United~\cite{megc24-1}& - & 0.1800 \\
          & He et al.~\cite{megc24-2}& - & 0.1900 \\
          & ME-TST~\cite{METST}  & 0.0802 & 0.2167 \\
          & ME-TST+  & \textbf{0.0912} & \textbf{0.2723} \\
    \bottomrule
    \end{tabular}%
  \label{tab:mes}%
\end{table}%

\begin{table}[t]
  \centering
  \caption{ME Recognition Evaluation on CAS(ME)$^3$.}
  \setlength{\tabcolsep}{15.0pt}
    \begin{tabular}{cccc}
    \toprule
    Type & Method & UF1$\uparrow$  & UAR$\uparrow$\\
    \midrule
    \multirow{4}[2]{*}{R} & STSTNet~\cite{ststnet} & 0.3795 & 0.3792 \\
          & RCN-A~\cite{RCN-A} & 0.3928 & 0.3893 \\
          & FeatRef~\cite{FR} & 0.3493 & 0.3413 \\
          & AlexNet~\cite{casme3} & 0.3001 & 0.2982 \\
    \midrule
    \multirow{4}[2]{*}{S \& R} & MEAN~\cite{MEAN} & 0.3894 & 0.4004 \\
          & SFAMNet~\cite{sfamnet} & 0.4462 & 0.4767 \\
          & ME-TST~\cite{METST} & 0.4754 & 0.4878 \\
          & ME-TST+  & \textbf{0.5317} & \textbf{0.5083} \\
    \bottomrule
    \end{tabular}%
  \label{tab:mer}%
\end{table}%

\begin{figure}[t]
  \centering
  \subfloat[CAS(ME)$^3$.]{\includegraphics[width=0.24\textwidth]{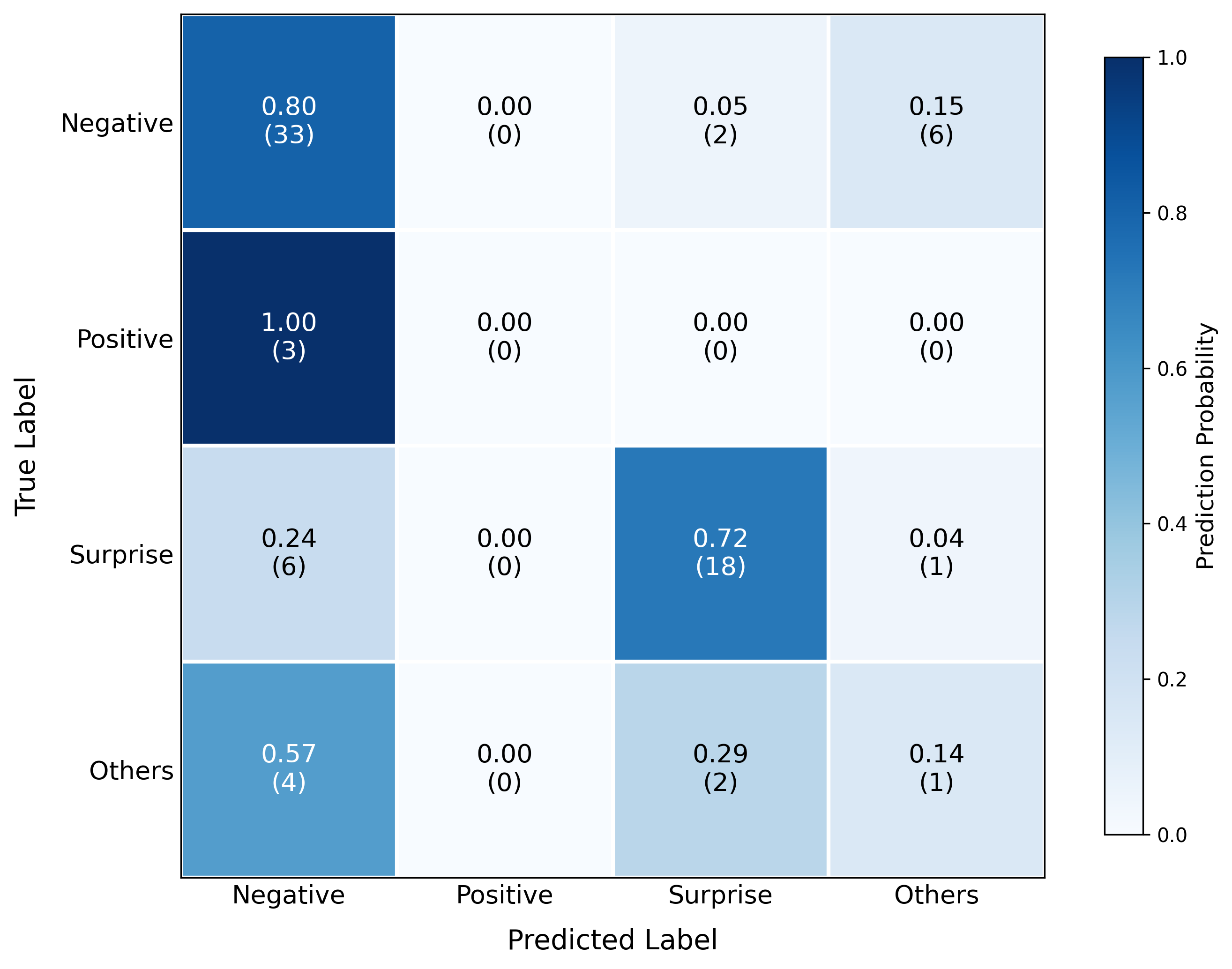}}
 \hfill 	
  \subfloat[SAMMLV.]{\includegraphics[width=0.24\textwidth]{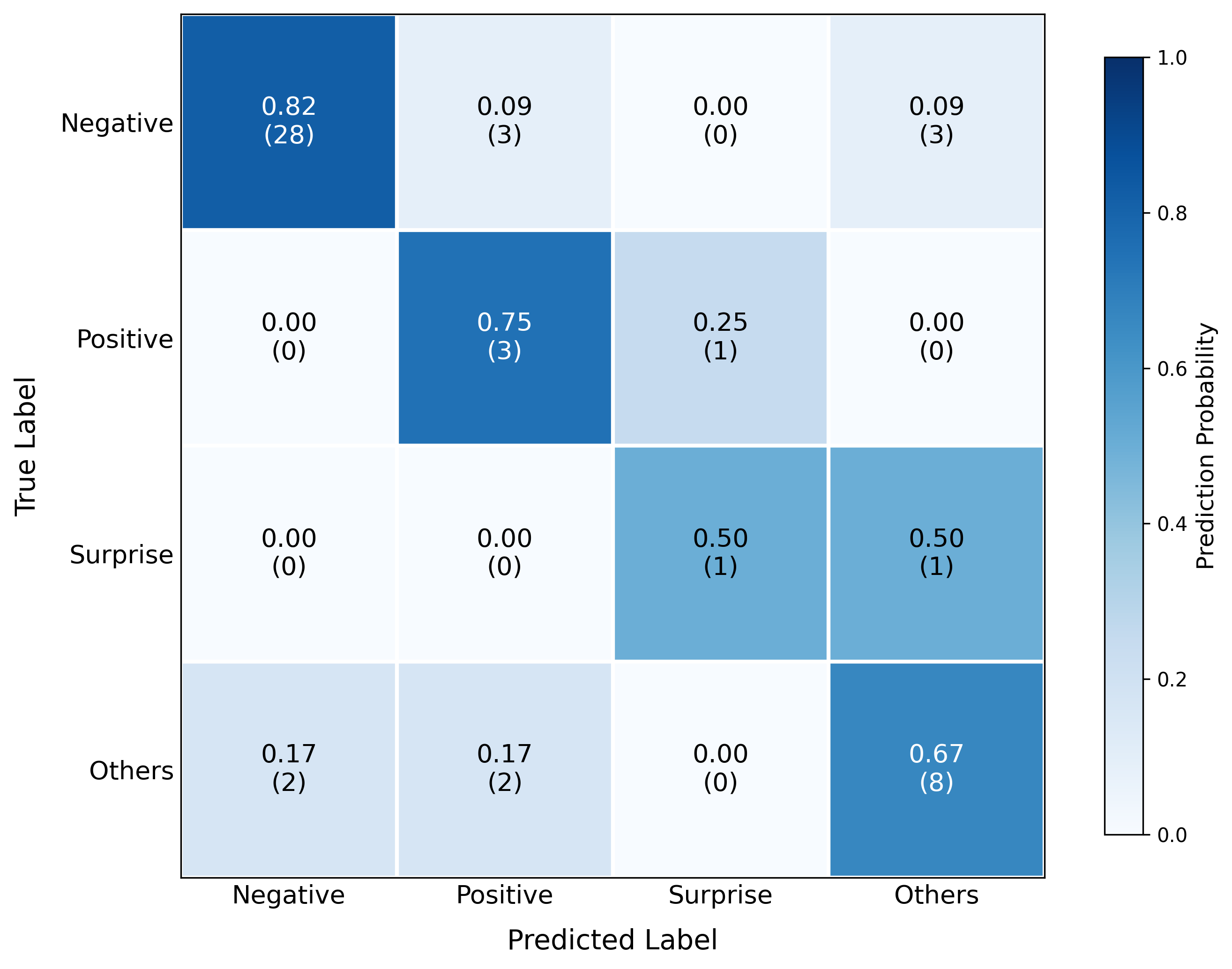}}
  \caption{Heatmap of ME Recognition Evaluation.}
  \label{fig:vis}
\end{figure}

\subsection{ME Spotting Evaluation}

ME spotting performance was evaluated according to the protocols established in~\cite{casme3,sfamnet}. We compared the proposed methods with several state-of-the-art approaches, including methods dedicated exclusively to spotting (Type S) and those addressing both spotting and recognition tasks simultaneously~(Type S\&R). As shown in Table~\ref{tab:mes}, ME-TST+ not only surpasses all existing methods in both spotting and recognition, but also achieves performance comparable to the state-of-the-art spotting-only approaches. This demonstrates the feasibility of utilizing the temporal state transition architecture for ME analysis in long videos. Furthermore, performing recognition concurrently with spotting not only does not compromise the spotting performance but also leads to its improvement by leveraging feature-level representation sharing and result-level synergy between the two tasks.

\begin{table*}[t]
  \caption{ME Analysis Evaluation. Analysis: STRS~$\uparrow$; Spotting / Recognition: F1-Score~$\uparrow$.}
  \centering
  \setlength{\tabcolsep}{10pt}
    \begin{tabular}{ccccccc}
    \toprule
    \multirow{2}[4]{*}{Method} & \multicolumn{3}{c}{CAS(ME)$^3$} & \multicolumn{3}{c}{SAMMLV} \\
    \cmidrule{2-7}   & Analysis & Spotting  & Recognition & Analysis & Spotting  & Recognition \\
    \midrule
    MEAN~\cite{MEAN} & 0.0100 & 0.0283 & 0.3532 & 0.0499 & 0.0949 & 0.5263 \\
    SFAMNet~\cite{sfamnet} & 0.0331 & 0.0716 & 0.4619 & -     & -     & - \\
    USTC-IAT-United~\cite{megc24-1}& -     & -     & - & 0.1100 & 0.1800  & 0.5800 \\
    He et al.~\cite{megc24-2}& -     & -     & - & 0.0900 & 0.1900  & 0.5100 \\
    ME-TST~\cite{METST}  & 0.0387 & 0.0802 & 0.4830 & 0.1356 & 0.2167 & 0.6259 \\
    ME-TST+  & \textbf{0.0487} & \textbf{0.0912} & \textbf{0.5338} & \textbf{0.1848} & \textbf{0.2723} & \textbf{0.6787} \\
    \bottomrule
    \end{tabular}%
  \label{tab:mea}%
\end{table*}%

\begin{table*}[t]
  \caption{Detailed Results of Result-Level Synergy Ablation.}
  \centering
    \begin{tabular}{ccccccccc}
    \toprule
          & \multicolumn{4}{c}{CAS(ME)$^3$} & \multicolumn{4}{c}{SAMMLV} \\
\cmidrule(lr){2-5} \cmidrule(){6-9}    
          & \multicolumn{2}{c}{w/o Result Synergy} & \multicolumn{2}{c}{w. Result Synergy} & \multicolumn{2}{c}{w/o Result Synergy} & \multicolumn{2}{c}{w. Result Synergy} \\
\cmidrule(lr){2-3} \cmidrule(r){4-5} \cmidrule(lr){6-7} \cmidrule(){8-9} 
    Metric & Spotting  & Recognition & Spotting  & Recognition & Spotting  & Recognition & Spotting  & Recognition \\
    \midrule
    Total & 858    & 72    & 858    & 69    & 159   & 41   & 159   & 40 \\
    TP    & 81     & 52    & 76     & 51    & 53    & 32   & 52    & 32 \\
    FP    & 912    & 13    & 733    & 11    & 184   & 5    & 171   & 4 \\
    FN    & 777    & 20    & 782    & 18    & 106   & 9    & 107   & 8 \\
    Precision & 0.0816 & 0.5519 & 0.0939 & 0.5619 & 0.2236 & 0.611 & 0.2332 & 0.6677 \\
    Recall & 0.0944 & 0.4921 & 0.0886 & 0.5083 & 0.3333 & 0.6833 & 0.3270 & 0.6912 \\
    F1-score & 0.0875 & 0.5202 & 0.0912 & 0.5338 & 0.2677 & 0.6452 & 0.2723 & 0.6787 \\
    STRS  & \multicolumn{2}{c}{0.0455} & \multicolumn{2}{c}{0.0487} & \multicolumn{2}{c}{0.1727} & \multicolumn{2}{c}{0.1848} \\
    \bottomrule
    \end{tabular}%
  \label{tab:arch1}%
\end{table*}%

\begin{table*}[t]
  \caption{Ablation Study. Analysis: STRS~$\uparrow$; Spotting / Recognition: F1-Score~$\uparrow$.}
  \centering
    \setlength{\tabcolsep}{6.5pt} 
    \begin{tabular}{cccccccccc}
    \toprule
    \multirow{1}[3]{*}{Multi-granularity} & \multirow{1}[3]{*}{Slowfast} & \multirow{1}[3]{*}{Feature-level} & \multirow{1}[3]{*}{Result-level} & \multicolumn{3}{c}{CAS(ME)$^3$} & \multicolumn{3}{c}{SAMMLV} \\
    \cmidrule(l){5-10}
       ROI & Mamba & Synergy & Synergy & Analy & Spot  & Recog & Analy & Spot  & Recog \\
    \midrule
    × & ×   & ×   & ×     & 0.0314 & 0.0753 & 0.4171 & 0.1321 & 0.2130 & 0.6202 \\
    × & ×   & \checkmark & \checkmark   & 0.0387 & 0.0802 & 0.4830 & 0.1356 & 0.2167 & 0.6259 \\
    ×  & \checkmark   & \checkmark   & ×     & 0.0336 & 0.0681 & 0.4941 & 0.0851 & 0.1971 & 0.4318 \\
    \checkmark & ×   & \checkmark   & ×     & 0.0354 & 0.0758 & 0.4670 & 0.1221 & 0.2513 & 0.4860 \\
    \checkmark & \checkmark   & ×   & ×     & 0.0406 & 0.0813 & 0.4985 & 0.1392 & 0.2400 & 0.5798 \\
    \checkmark & \checkmark   & \checkmark &  ×  & 0.0455 & 0.0875 & 0.5202 & 0.1727 & 0.2677 & 0.6452 \\
    \checkmark & \checkmark   & \checkmark &  \checkmark    & \textbf{0.0487} & \textbf{0.0912} & \textbf{0.5338} & \textbf{0.1848} & \textbf{0.2723} & \textbf{0.6787} \\
    \bottomrule
    \end{tabular}%
  \label{tab:ablation}%
\end{table*}%

\begin{figure*}[htbp]
  \centering  
  \includegraphics[width=0.78\linewidth]{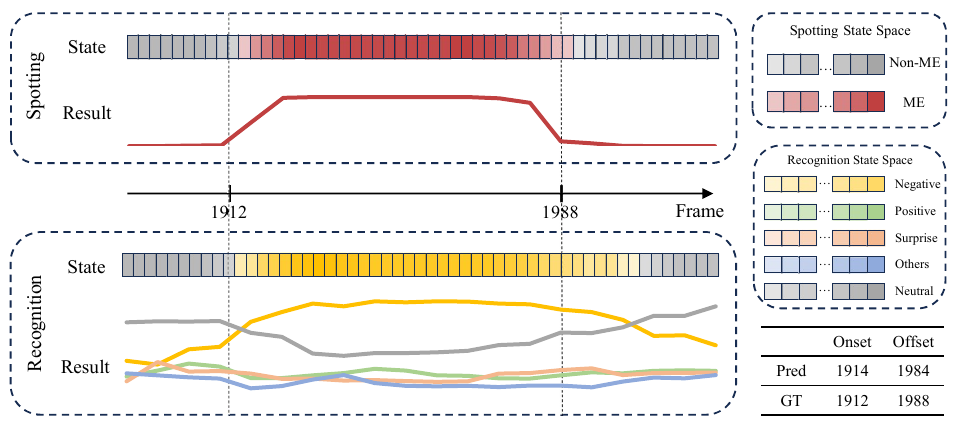}  
  \caption{Visualization of a sample result from the SAMMLV dataset using ME-TST+.}
  \label{fig:vis1}
\end{figure*}

\begin{figure*}[htbp]
  \centering  
  \includegraphics[width=0.78\linewidth]{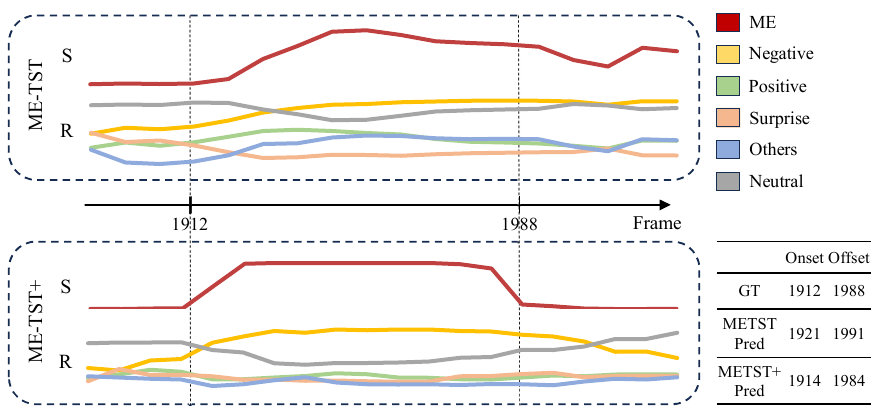}  
  \caption{Comparative Visualization of ME-TST and ME-TST+. 'S' denotes spotting and 'R' denotes recognition.}
  \label{fig:vis2}
\end{figure*}

\subsection{ME Recognition Evaluation}
We conducted a comprehensive evaluation of ME recognition performance on the CAS(ME)$^3$ dataset, following the protocol outlined in~\cite{casme3,sfamnet}, in which MEs are categorized into four classes: Negative, Positive, Surprise, and Others. As shown in Table~\ref{tab:mer}, ME-TST+ achieves the highest scores, with a UF1 of 0.5151 and a UAR of 0.4930. Compared with previous methods, ME-TST+ not only outperforms methods that simultaneously spot and recognize~(S\&R) MEs, but also surpasses methods that perform recognition only~(R), demonstrating consistent and substantial improvements. These results illustrate the effectiveness and robustness of our unified approach for real-world ME analysis.

Although recognition-only~(R) methods and spot-and-recognize (S\&R) frameworks are compared together, their evaluation approaches differ substantially. On the one hand, recognition-only methods are evaluated using all ground-truth intervals. In contrast, S\&R frameworks calculate recognition results only for the true positive intervals identified during the spotting stage. Therefore, intervals missed during spotting—often those representing more challenging cases—are omitted from the recognition evaluation. On the other hand, recognition-only approaches have access to precise ME onset and apex annotations, which define strict temporal boundaries. In contrast, ME analysis in real-world environments requires direct spotting and recognition from long video sequences, where access to ground-truth interval information is unavailable. This substantially increases the task difficulty compared to recognition-only protocols. Since intervals are considered true positives as long as the IoU is greater than 0.5, the detected onset and apex frames often exhibit minor deviations, and sometimes even large discrepancies. The proposed temporal state transition recognition branch not only effectively alleviates the limitations imposed on the recognition task by imperfect spotting results, but also turns these limitations into advantages: it facilitates model optimization and decision-making for the spotting task itself. In this way, our approach transforms a potential weakness into a strength, enabling mutual enhancement between the recognition and spotting components.

Fig.~\ref{fig:vis} presents confusion matrix heatmaps for both the CAS(ME)$^3$ and SAMMLV datasets. The pronounced diagonal entries indicate reliable classification across emotional categories, notably for the Negative and Surprise classes. However, recognition of the Positive category in the CAS(ME)$^3$ dataset remains challenging. None of the three spotted positive samples was correctly classified. This can largely be attributed to the severe class imbalance inherent in ME datasets, which remains a significant challenge and an important direction for future research.

\subsection{ME Analysis Evaluation}
We conducted ME analysis following established protocols outlined in~\cite{sfamnet,MEAN}, where all MEs (Negative, Positive, Surprise, and Others) are considered for temporal spotting, while only the first three are used for recognition. As summarized in Table~\ref{tab:mea}, our proposed video-level regression framework consistently outperforms previous approaches across all primary metrics—STRS, spotting F1-score, and recognition F1-score—demonstrating clear advantages over conventional window-based classification methods.

A more detailed breakdown is provided in Table~\ref{tab:arch1}, presenting both the instance counts and the performance statistics for spotting and recognition tasks. Notably, our method achieves stable and superior results on both the CAS(ME)$^3$ and SAMMLV datasets, with improvements observed across all key metrics. Of particular note, precision and recall remain closely matched on both datasets, reflecting a balanced capacity for accurate temporal spotting and subsequent recognition.

Furthermore, it can be observed that the overall performance on CAS(ME)$^3$ is relatively lower, primarily due to limitations in the spotting results. This may be attributed to the more challenging recording conditions of the dataset. Nevertheless, the results presented in both tables demonstrate that our proposed approach exhibits enhanced robustness and generalization ability compared to previous methods, across both challenging and relatively tractable ME analysis scenarios.

\subsection{Ablation Study}
We conducted ablation studies on the CAS(ME)$^3$ and SAMMLV datasets to evaluate the individual contributions of multi-granularity ROI, slowfast mamba, and synergy strategy.

\textbf{Impact of Multi-granularity ROI.} In ME-TST, only the most salient ROI optical flow is extracted at a relatively coarse granularity, which limits its ability to adapt to the diverse manifestations of MEs caused by individual differences in intensity and movement patterns. As shown in the third and sixth rows of Table~\ref{tab:ablation}, the introduction of multi-granularity ROI extraction partially alleviates the information loss associated with treating the video analysis task solely as a temporal sequence analysis problem.

\textbf{Impact of SlowFast Mamba.} As shown in the fourth and sixth rows of Table~\ref{tab:ablation}, the slowfast mamba architecture effectively leverages temporal information via its dual-pathway design, leading to improved performance. This effect is particularly pronounced when integrated with multi-granularity ROI features, which provide richer and more comprehensive information.

\textbf{Impact of Synergy Strategy.} 
The synergy strategy encompasses both feature-level representation sharing and result-level collaboration between the two tasks, each of which was isolated in ablation studies for independent evaluation. In the feature-level synergy setting, the loss weight for the neutral category was set to zero; whereas, in the result-level synergy scenario, outputs were generated by sequentially performing spotting and then recognition. As shown in Table~\ref{tab:ablation}, the comparison of the fifth, sixth, and seventh rows demonstrates that the proposed synergy strategy effectively enhances the overall performance at both levels. This indicates that the synergy framework successfully leverages the complementary strengths of both spotting and recognition, resulting in a joint performance that surpasses the sum of individual task contributions.
In addition, Table~\ref{tab:arch1} presents detailed results of the result-level synergy ablation. It can be observed that, while there is a slight sacrifice in recall, there is a substantial improvement in precision, indicating the feasibility of recognition-assisted spotting.

\begin{figure*}[htbp]
  \centering
  \subfloat[]{\includegraphics[width=0.467\textwidth]{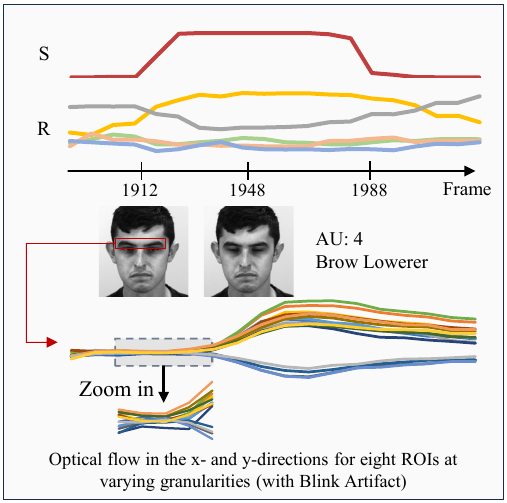}}
  \hspace{0.04\textwidth} 
  \subfloat[]{\includegraphics[width=0.47\textwidth]{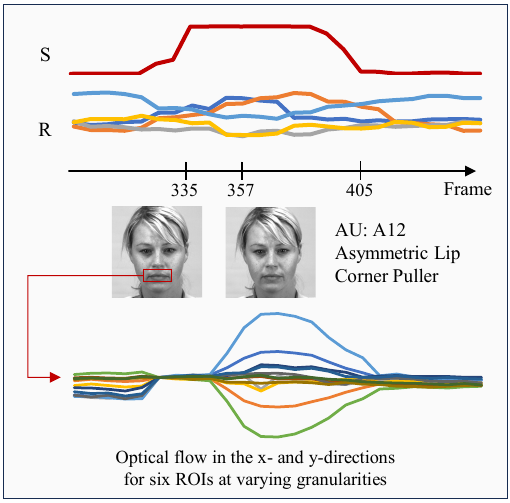}}
  \caption{Visualization of temporal probability outputs (for both spotting and recognition) and the corresponding key ROI optical flow. 'S' denotes spotting and 'R' denotes recognition.}
  \label{fig:vis3}
\end{figure*}

\subsection{Visualization}
We present a visualization example from the SAMMLV dataset in Fig.~\ref{fig:vis1}. At critical temporal points—onset, apex, and offset—both the spotting and recognition outputs display notable responses, demonstrating the effectiveness of our approach in capturing ME states and supporting the feasibility of the proposed temporal state transition framework. Notably, the negative emotion probability curve (yellow) in the recognition output closely aligns with the spotting probability results, suggesting that the recognition branch inherently possesses certain localization capabilities.

Fig.~\ref{fig:vis2} presents a comparative visualization of ME-TST and ME-TST+. Compared to ME-TST, ME-TST+, which leverages multi-granularity ROI and the slowfast Mamba to fully exploit spatiotemporal information, produces more stable probabilities regarding the occurrence of MEs and achieves higher discriminability among similar categories, as well as more accurate localization of onset and offset points. In addition, both methods effectively capture the temporal dynamics of MEs, which demonstrates the feasibility and advantages of the temporal state transition framework.

Fig.~\ref{fig:vis3} presents the visualization of temporal probability results for spotting and recognition, as well as key ROI optical flow. The emotions in examples a and b correspond to anger and happiness, respectively. It can be observed that the network has successfully learned the correspondence between the motion of key ROIs and the occurrence and category of MEs to a certain extent. Notably, in example a, a blink occurs concurrently with the ME and produces motion significantly larger than that of the ME itself. The network effectively captures the onset and offset of the ME under this challenging condition, demonstrating its robustness to noise.
\label{sec/5_conclusion}
\section{Conclusion}

In this paper, we propose two novel temporal state transition architectures for ME analysis, namely ME-TST and ME-TST+, which reformulate the traditional window-level hard classification as a video-level regression. This paradigm better aligns with the intrinsic characteristics of MEs, and its linear complexity allows for efficient processing of long video sequences. Furthermore, by employing multi-granularity ROIs and the slowfast mamba network, our method can comprehensively capture spatiotemporal information, achieving state-of-the-art performance across multiple evaluation benchmarks. Additionally, we exploit the inherent relationship between spotting and recognition by introducing a synergy strategy at both the feature and result levels, thereby achieving a synergistic performance improvement.
In future work, we plan to further explore the potential of this architecture by extending its analytical capabilities from MEs to MaEs, rather than restricting its application solely to ME analysis.

\bibliographystyle{IEEEtran}
\bibliography{main}

\vfill

\end{document}